\def\year{2020}\relax
\documentclass[letterpaper]{article} 
\usepackage{aaai20}
\usepackage{times} 
\usepackage{helvet}
\usepackage{courier}
\usepackage[hyphens]{url}
\usepackage{graphicx}
\usepackage{graphicx}
\usepackage{multirow}
\usepackage{colortbl} 
\usepackage{arydshln}
\usepackage{amsmath,amssymb,amsfonts}
\usepackage{subcaption}
\usepackage{booktabs}
\newcommand {\citet}[1]{\citeauthor{#1} \shortcite{#1}}
\newcommand {\citep}{\cite}
\title{Gated Convolutional Networks with Hybrid Connectivity for Image Classification}

\author{Chuanguang Yang,\textsuperscript{\rm 1,2} Zhulin An,\textsuperscript{\rm 1}\thanks{Corresponding author} Hui Zhu,\textsuperscript{\rm 1,2} Xiaolong Hu,\textsuperscript{\rm 1,2}  \\ \bf \Large{Kun Zhang,\textsuperscript{\rm 1,2} Kaiqiang Xu,\textsuperscript{\rm 1,2} Chao Li,\textsuperscript{\rm 1} Yongjun Xu\textsuperscript{\rm 1}}\\ 
\textsuperscript{\rm 1}Institute of Computing Technology, Chinese Academy of Sciences, Beijing, China\\ 
\textsuperscript{\rm 2}University of Chinese Academy of Sciences, Beijing, China\\
\{yangchuanguang, anzhulin, zhuhui, huxiaolong18g, zhangkun17g, xukaiqiang, lichao, xyj\}@ict.ac.cn 
}

\begin{document}
 	
 	\maketitle
 	
 	\begin{abstract}
 		We propose a simple yet effective method to reduce the redundancy of DenseNet by substantially decreasing the number of stacked modules by replacing the original bottleneck by our SMG module, which is augmented by local residual. Furthermore, SMG module is equipped with an efficient two-stage pipeline, which aims to DenseNet-like architectures that need to integrate all previous outputs, i.e., squeezing the incoming informative but redundant features gradually by hierarchical convolutions as a hourglass shape and then exciting it by multi-kernel depthwise convolutions, the output of which would be compact and hold more informative multi-scale features. We further develop a forget and an update gate by introducing the popular attention modules to implement the effective fusion instead of a simple addition between reused and new features.  Due to the \textbf{H}ybrid \textbf{C}onnectivity (nested combination of global dense and local residual) and \textbf{G}ated mechanisms, we called our network as the \textbf{HCGNet}. Experimental results on CIFAR and ImageNet datasets show that HCGNet is more prominently efficient than DenseNet, and can also significantly outperform state-of-the-art networks with less complexity. Moreover, HCGNet also shows the remarkable interpretability and robustness by network dissection and adversarial defense, respectively. On MS-COCO, HCGNet can consistently learn better features than popular backbones.
 	\end{abstract}
 	\section{Introduction}
\begin{figure}[tbp]  
 		\centering  
 		\includegraphics[width=.95\columnwidth]{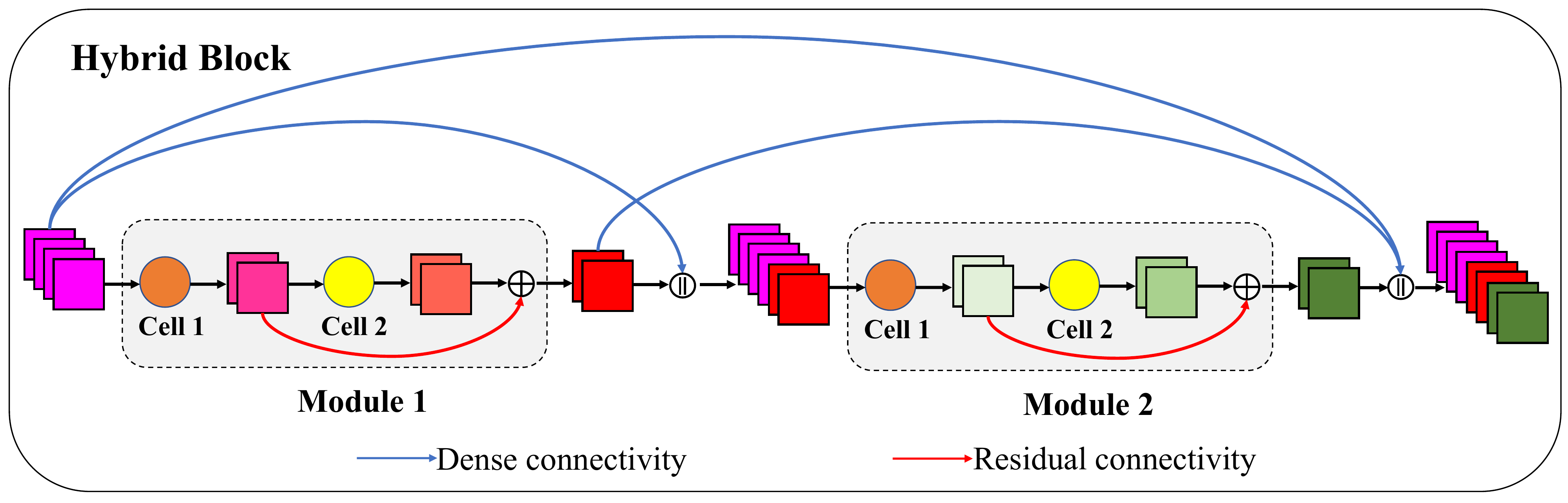}
 		\caption{The diagram of a hybrid block including $n=2$ modules, where $n\geqslant  2$. The symbol "$+$" and "$\parallel $" denote element-wise addition and channel-wise concatenation among multiple feature maps, respectively.}  
 		\label{fig:figure1}
 	\end{figure}

 	Deep convolutional neural networks (CNNs) are becoming more and more efficient in parameter and computation without sacrificing the performance owing to novel architectures design. ResNet \cite{he2016deep} introduces the residual connectivity to implement the addition of the input and output features for each micro-block. DenseNet \cite{huang2017densely} holds the dense connectivity by changing skip connections from addition to concatenation. Both of their feature aggregation connectivities can not only encourage feature reuse, but also ease the training problems. For a detailed comparison, dense connectivity is more effect for feature exploitation and exploration but exists a certain redundancy, while residual connectivity contributes to efficient feature reuse by parameter sharing mechanism and thus leads to low redundancy, but lacks the capability of feature preservation and exploration. To enjoy their advantages and avoid inherent limitations, many networks combine them to build a more effective aggregation topology, such as DPN \cite{chen2017dual}, MixNet \cite{wang2018mixed} and AOGNet \cite{AOGNets}. Different from them, we develop a hybrid connectivity (Figure \ref{fig:figure1}) with nested aggregation that facilitates feature flow by dense connectivity for global channel-wise concatenation of outputs produced by all precedent modules (blue links in Figure \ref{fig:figure1}) and residual connectivity for local element-wise addition within the module (red links in Figure \ref{fig:figure1}).
 	
 Our main motivation for this pattern design originates from reducing the redundancy of dense connectivity. As the depth of network linearly increases, the number of skip connections and required parameters grow by a rate of $O(n^{2})$, where $n$ denotes the number of stacked modules under dense connectivity. Meanwhile, early superfluous features which have few contributions are transferred quadratically to subsequent modules. So one simple method to reduce redundancy is to decrease the number of modules directly, but it can attenuate the representational power of features and then deteriorate the performance. Thus we develop a novel module by embedding the residual connectivity to assist feature learning within the local module. Experimentally, the number of our proposed modules under dense connectivity can be quite fewer than that of classical modules in the dense block but without sacrificing the performance. 
 
 For further adaptation with hybrid connectivity, we instantiate the basic module that includes a \textbf{squeeze cell (cell 1 in Figure \ref{fig:figure1})} for transforming the input to a compact feature map, and a \textbf{multi-scale excitation cell (cell 2 in Figure \ref{fig:figure1})} to further extract multi-scale features by multi-kernel convolutions. It is widely known that convolution builds pixel relationship in a local neighborhood, which leads to ineffective modeling of long-range dependency. To fully address this issue, we develop an update gate to model the global context features from more informative multi-scale features. Moreover, we locate a forget gate on the residual connection to capture channel-wise dependency for decaying the reused features produced by cell 1. Finally, global context features are added to the reused feature map of each spatial position to form the output, which can not only promote effective feature exploration but also retain the capability of feature re-exploitation to some extent. Moreover, both forget gate and update gate are lightweight and general plug-ins, which can be integrated into any CNNs with negligible overheads.
 
 We perform extensive experiments across the three highly competitive image classification datasets: CIFAR-10/100 \cite{krizhevsky2009learning}, and ImageNet (ILSVRC 2012) \cite{deng2009imagenet}. On CIFAR datasets, HCGNets outperform state-of-the-art both human-designed and auto-searched networks but only requiring extremely fewer parameters, e.g., HCGNet-A3 obtains the better result than the most competitive NASNet-A \cite{zoph2018learning} with $4.5\times$ fewer parameters. On ImageNet datasets, it also consistently obtains the best accuracy, interpretability, robustness based on classification and transferability to object detection as well as segmentation among the widely used networks with least complexity, e.g., HCGNet-B outperforms previous SOTA AOGNet across a broad range of tasks with similar complexity.

 	\section{Related Work}
 	
 	\subsubsection{Improvements of ResNet and DenseNet.}ResNeXt \cite{xie2017aggregated} outperforms ResNet with less overheads since it adopts 3$\times $3 group convolutions in residual blocks. Afterwards, group convolutions become popular in efficient CNNs design due to the properties of lower parameter and computational cost, including our HCGNets. Wide ResNet \cite{Wide} show that increasing width while decreasing depth of residual networks can surpass very deep counterparts, meanwhile tackling the problems of slow training and weakened feature reuse. By representing the multi-scale features and widening the receptive fields (RF) within the residual block, Res2Net \cite{gao2019res2net} outperforms the other backbones across a broad range of tasks. Multi-scale information has been widely demonstrated a effective way to improve the performance, our HCGNet also constructs the multi-scale features by multi-branch convolutions.  
 	
 	It is widely known that DenseNet has a certain redundancy, thus a typical practice is sparsification. Log-DenseNet \cite{hu2017log} and SparseNet \cite{zhu2018sparsely} regularly conduct a sparse rather than full aggregation of all previous outputs, which change the number of connections from linear to be logarithmic in the overall topology. Learned group convolutions are adopted in CondenseNet \cite{huang2018condensenet} to automatically prune unimportant channels for the incoming feature map based on channel-wise L1-norm. However, excessive sparsification affects the superiority of collective learning. Thus we only decrease the number of modules under dense connectivity to reduce redundancy, which is empirically more effective than sparsification.
 	
 	\subsubsection{Combinations of ResNet and DenseNet.} To enjoy the advantages and avoid drawbacks of both two connectivities, many combinations have proposed. DPN adopts dual path architectures, which can facilitate effective feature reuse by residual path and  feature exploration by dense path in parallel. MixNet blends two connectivities to implement feature aggregation with more flexible positions and sizes, further ResNet, DenseNet and DPN can be treated as particular cases of MixNet. Recently proposed AOGNet utilizes AND-OR Grammar to generate CNNs by parsing feature map as a sentence, where AND-node denotes channel-wise concatenation and OR-node denotes element-wise addition. It demonstrates that the compositional and hierarchical aggregation in AOGNet is more effective than cascade-based way in DPN. Moreover, addition and concatenation as the meta-operations are also widely applied in the field of neural architecture search, such as NASNet, PNASNet \cite{liu2018progressive} and AmoebaNet \cite{real2019regularized}. Extensive experiments indicate that the nested way for feature aggregation in our HCGNets perform the best. 
 	
 	\subsubsection{Attention Mechanisms}Attention has been widely applied in computer vision, e.g., image classification \cite{wang2017residual}. SENet \cite{hu2018squeeze} introduces a lightweight gate to capture channel-wise dependencies for rescaling channel features. SKNet \cite{li2019selective} further employs a dynamic kernel selection attention for weighted multi-scale features fusion, which is inspired by InceptionNets \cite{szegedy2017inception}. Beyond channel, CBAM \cite{woo2018cbam} also constructs a spatial attention map to recalibrate spatial features. To capture long-range dependency, GCNet \cite{cao2019gcnet} simplifies non-local block \cite{wang2018non} to implement query-independent context modeling based on single branch information. Different from them in roles or mechanisms, we build a forget gate to capture channel-wise dependency for decaying the reused features, while an update gate fully models the global context features from multi-scale information.
 	\section{Revisiting ResNet and DenseNet}
 	We revisit the classical ResNet and DenseNet with their individual residual connectivity and dense connectivity, and further investigate their mechanisms of parameter sharing and feature learning. Finally, we analyse the overall efficiency of ResNet and DenseNet.    
 	
 	\subsection{Parameter Sharing}
 	Intuitively, residual connectivity implicitly accompanies a parameter sharing mechanism between the reused and newly extracted features by processing their mixed features. We now formally describe why the parameter sharing mechanism can take place in residual connectivity but not in dense connectivity. Concretely, we use $\mathcal{F}$ to denote the bottleneck unit. Consider the input feature map $\mathbf{x}_{l-1}\in \mathbb{R}^{H\times W\times C}$ to the $l$-th residual block, corresponding formula is as follows:
 	\begin{equation}
 	\label{trad}
 	\mathbf{x}_{l}=\mathbf{x}_{l-1}+\mathcal{F}_{l}(\mathbf{x}_{l-1};W_{l})=\mathbf{x}_{l-1}+\mathbf{\tilde{x}}_{l}
 	\end{equation}
 	Where $\mathbf{x}_{l-1}$ can be considered as the reused feature map, $W_{l}$ and $\mathcal{F}_{l}(\mathbf{x}_{l-1};W_{l})$ refer to convolutional weights and newly extracted feature map, respectively. $\mathbf{\tilde{x}}_{l}\in \mathbb{R}^{H\times W\times C}$ represents $\mathcal{F}_{l}(\mathbf{x}_{l-1};W_{l})$ for simplicity. Afterwards, $\mathbf{x}_{l}$ becomes a new input for the next residual block to proceed the transformation:
 	\begin{align}
 	\label{res}
 	\mathbf{x}_{l+1}&=\mathbf{x}_{l}+\mathcal{F}_{l+1}(\mathbf{x}_{l};W_{l+1}) \notag\\
 	&=\mathbf{x}_{l}+\mathcal{F}_{l+1}(\mathbf{x}_{l-1}+\mathbf{\tilde{x}}_{l};W_{l+1})
 	\end{align}
 	In the $l+1$-th residual block, $\mathbf{x}_{l-1}$ and $\mathbf{\tilde{x}}_{l}$ are shared with the same $W_{l+1}$ and operations. Similar analysis about dense connectivity is exhibited as follows. Output of the $l$-th module under dense connectivity can be regarded as the concatenation of input $\mathbf{x}_{l-1}\in \mathbb{R}^{H\times W\times C}$ and newly extracted feature map $\mathbf{\tilde{x}}_{l}\in \mathbb{R}^{H\times W\times \tilde{C}}$ along the channels:
 	\begin{align}
 	\label{dense}
 	\mathbf{x}_{l}&=\mathbf{x}_{l-1}\parallel  \mathcal{F}_{l}(\mathbf{x}_{l-1};W_{l})
 	=\mathbf{x}_{l-1}\parallel \mathbf{\tilde{x}}_{l}
 	\end{align}
 	Then, the next module receives $\mathbf{x}_{l}\in \mathbb{R}^{H\times W\times (C+\tilde{C})}$ and conducts the following transformation:
 	\begin{align}
 	\label{denss}
 	\mathbf{x}_{l+1}&=\mathbf{x}_{l}\parallel \mathcal{F}_{l+1}(\mathbf{x}_{l};W_{l+1}) \notag\\
 	&=\mathbf{x}_{l}\parallel \mathcal{F}_{l+1}(
 	\mathbf{x}_{l-1}\parallel \mathbf{\tilde{x}}_{l}	
 	;W_{l+1})
 	\end{align}
 	
 	In the $l+1$-th dense block, $\mathbf{x}_{l-1}$ and $\mathbf{\tilde{x}}_{l}$ are not shared with the same $W_{l+1}$ because of the different locations of feature space between reused and newly extracted feature maps.
 	
 	\subsection{Feature Learning}
 	The final output of residual block is the element-wise addition of input  and newly extracted feature maps. This addition pattern facilitates efficient feature reuse without increasing the size of feature map thus reducing parameter redundancy. But one potential fact is that too many aggregations by addition may collapse the feature representation and thus impede the information flow, hence some early informative features may be lost inevitably. Moreover, parameter sharing mechanism may damage the capability of exploring new features.
 	
 	Subsequently proposed DenseNet develops a global dense connectivity, where the output feature map of each preceding module flows to the all subsequent modules directly. Different from the element-wise addition, input and newly extracted feature maps are combined by concatenation along the channels. Thus dense connectivity can transfer the early feature maps to later modules, which preserves the all preceding feature information and facilitate the full exploitation of existing features. Moreover, various modules with different weights conduct a collective learning for the same features, which can promote effective feature exploration.
 	\subsection{Overall Efficiency} 	
 	It is widely known that DenseNet-100 with 0.8M parameters slightly outperforms ResNet-1001 with 10.2M parameters on CIFAR10 dataset. The explicit parameter gap is that DenseNet-100 is quite shallower than ResNet-1001 due to the more effective feature exploitation and exploration capabilities produced by collective learning, while ResNet mainly depends on increasing depth to improve the representational power of features. Empirically, DenseNet can also have extremely few number of filters in each convolutional layer due to the collective learning mechanism that further improve the efficiency. However, one potential weakness of dense connectivity is the redundancy of repeated extraction with the same features. In this case, early features flow to all subsequent layers, even if they have few contributions. By contrast, residual connectivity has a relatively low redundancy due to the parameter sharing mechanism.
 	
 	\section{Networks Architecture}
 	\subsection{Hybrid Connectivity Pattern}
 	We develop a hybrid connectivity pattern, which can enjoy the effective feature learning and few filters of each module from global dense connectivity as well as efficient feature reuse by parameter sharing from local residual connectivity. Figure \ref{fig:figure1} illustrates this pattern within the hybrid block schematically. Note that hybrid connectivity pattern exists in the hybrid block which consists of $n$ ($n\geqslant 2$) modules. To match the definition of growth rate in DenseNet, each module produces one feature map with $k$ channels. The basic module consists of successive two cells, which we call them as cell 1 and cell 2, respectively. Globally, input of each module is a concatenation of all feature-maps produced by preceding modules and transferred by dense connectivity. Locally, residual connectivity provides a shortcut that allows the output of cell 1 bypassing cell 2 and then being added to the new features generated by cell 2 to form the output.
 	
 	\subsection{Instantiation of Basic Module}
 	To orchestrate our hybrid connectivity, we design a basic SMG module which includes a \textbf{Squeeze cell} (cell 1), a \textbf{Multi-scale excitation cell} (cell 2) and \textbf{Gate mechanisms}. Unless specified otherwise, each convolution is bound a pre-activation, which refers to the three consecutive operations: batch normalization (BN)-rectified linear unit (ReLU)-Conv.
 
 	\begin{figure*}
 		\centering 
 		
 		\begin{subfigure}[t]{0.30\textwidth}
 			\centering
 			\includegraphics[width=\textwidth]{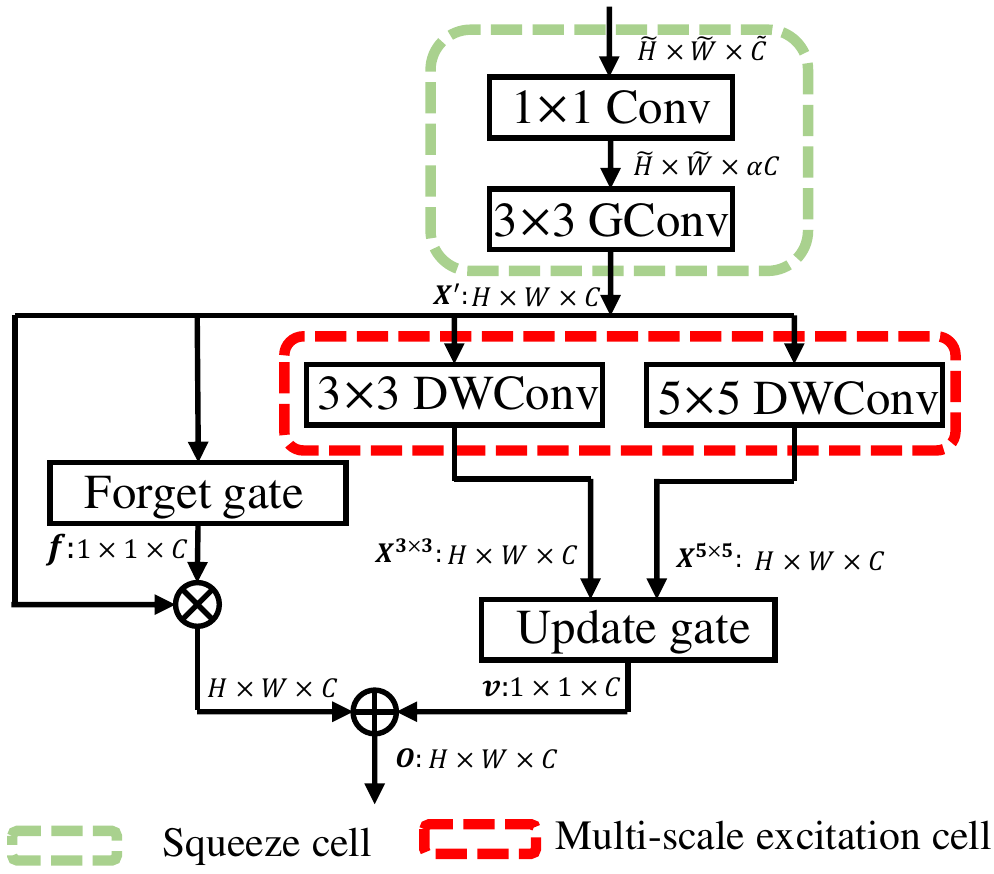}
 			\caption{SMG Module}
 			\label{SMG}
 		\end{subfigure}
 		\begin{subfigure}[t]{0.45\textwidth}
 			\centering
 			\includegraphics[width=\textwidth]{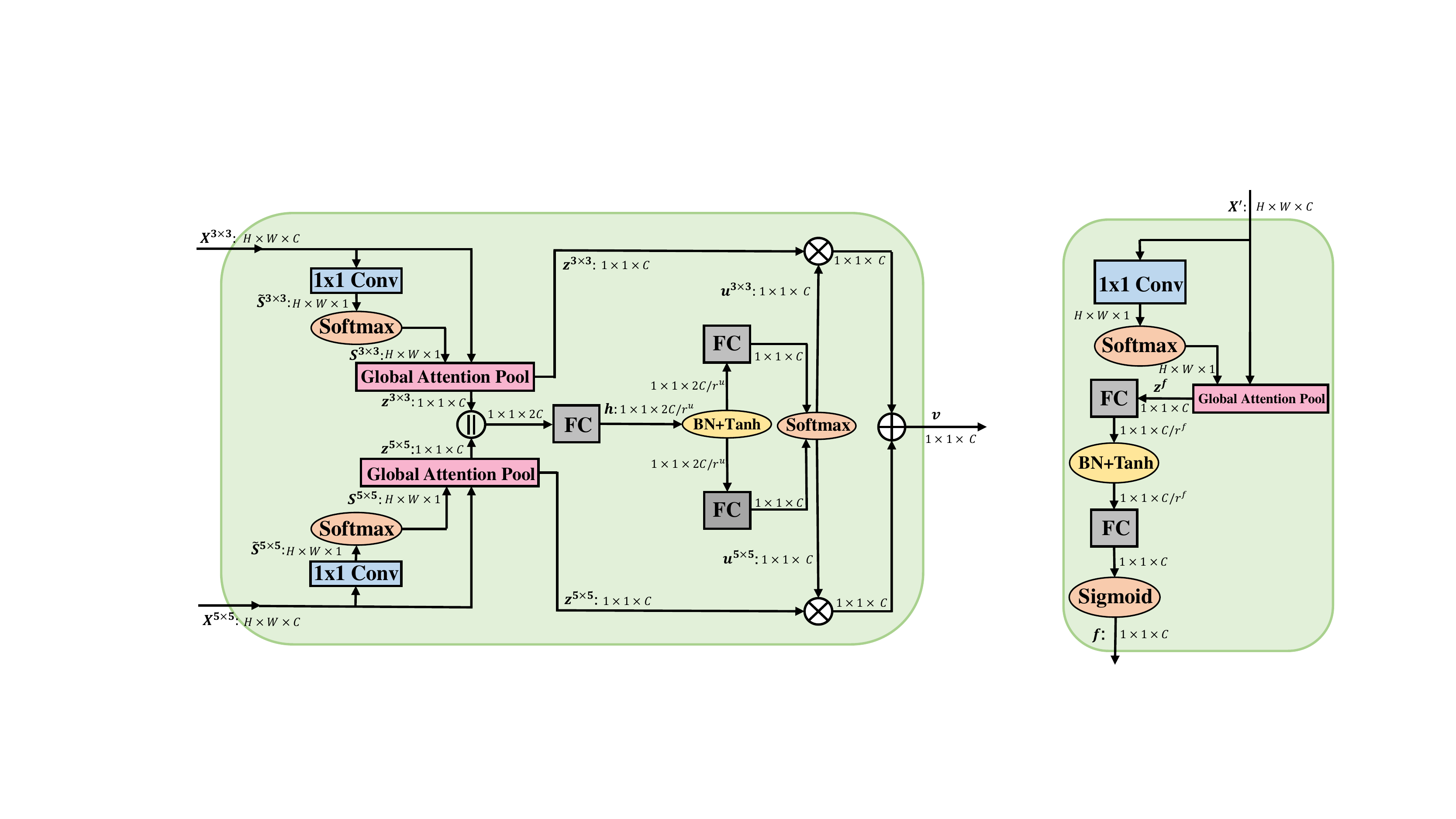}
 			\caption{Update Gate}
 			\label{Update}
 		\end{subfigure}
 	\begin{subfigure}[t]{0.16\textwidth}
 		\centering
 		\includegraphics[width=\textwidth]{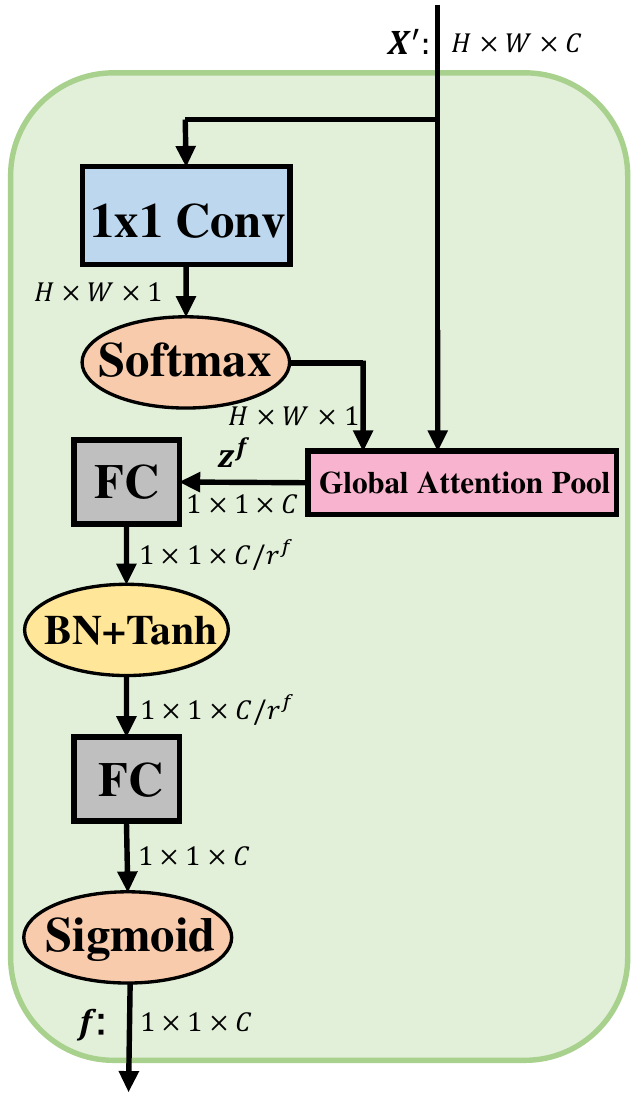}
 		\caption{Forget Gate}
 		\label{Forget}
 	\end{subfigure}

 		\label{Gate} 
 		\caption{Illustrations of SMG module, update gate and forget gate. In all figures, $\bigoplus$ and $\bigotimes$ denote broadcast element-wise addition and multiplication, respectively. We employ feature dimensions to describe the flow of feature maps for better understanding. Note that spatial size $\tilde{H}\times \tilde{W}= H\times W$ when default stride $S=1$ of 3$\times$3 GConv in Figure \ref{SMG}.} 
 	\end{figure*} 
 	
 	\subsubsection{Squeeze Cell.} This cell which locates at the beginning of SMG module is responsible for generating the compact feature map from input to improve parameter and computational efficiency for subsequent processing. 1$\times$1 convolution is firstly adopted for changing the number of input channels $\tilde{C}$ to $\left \lfloor \alpha\cdot C \right \rfloor$, where $\alpha>0$ can be reckoned as a width multiplier which is mostly used to reduce the number of channels, i.e.  $\tilde{C}>\left \lfloor \alpha\cdot C \right \rfloor$, and $C$ denotes the number of final output channels of squeeze cell. Then, 3$\times$3 group convolution (GConv) with $g$ groups proceeds to squeeze the features by reducing the number of channels from $\left \lfloor \alpha\cdot C \right \rfloor$ to $C$, where $C$ needs to be divisible by $g$. Moreover, it can also play a down-sampling by 3$\times$3 kernel with stride $S=2$.
 	\subsubsection{Multi-scale Excitation Cell.} Squeezed feature map enters this cell for multi-scale excitation by multi-branch convolutions with different kernel sizes. Note that the costs of parameter and computation are extremely cheap because of the few input channels, and the size of feature map throughout this cell is unchanged. To further improve efficiency, we adopt 3$\times$3 and 5$\times$5 depthwise convolutions (DWConv) with 1 and 2 paddings, respectively. Moreover, dilation convolution \cite{DBLP} with a kernel size of 3$\times$3 and a dilation size of 2 is used to approximate 5$\times$5 kernel for better trade-off between efficiency and performance. The output of this cell is two-branch feature maps produced by 3$\times$3 and 5$\times$5 DWConvs, respectively.

 	\subsubsection{Update Gate.}To capture long-range dependency , we utilize update gate to model the global context features from multi-scale information. Figure \ref{Update} shows the overall details about the update gate, which can be sequentially summarized for 3 stages: spatial attention, pooling and channel attention. 
 	
 	\textbf{\emph{spatial attention and pooling:}} We perform a global context modeling for calculating spatial-wise weights of each position. For the given feature map $\mathbf{X}^{3\times 3}\in \mathbb{R}^{H\times W\times C}$, a 1$\times$1 convolutional filter shrinks it along channel dimensions to a spatial attention map $\tilde{\mathbf{S}}^{3\times 3}\in \mathbb{R}^{H\times W\times 1}$, an then a softmax function normalizes it to obtain the final spatial attention map $\mathbf{S}^{3\times 3}\in \mathbb{R}^{H\times W\times 1}$, each element of which is as follows:
 	\begin{align}
 	\label{spatial}
 	\mathbf{S}^{3\times 3}_{i,j,1}=\frac{e^{\tilde{\mathbf{S}}^{3\times 3}_{i,j,1}}}{\sum_{x=1}^{H}\sum_{y=1}^{W}e^{\tilde{\mathbf{S}}^{3\times 3}_{x,y,1}}}
 	\end{align}
 	We employ global attention pooling via weighted averaging with $\mathbf{S}^{3\times 3}$ to shrink the global spatial information and generate the global context feature map $\mathbf{z}^{3\times 3}\in \mathbb{R}^{1\times 1\times C}$. The $c$-th channel of $\mathbf{z}^{3\times 3}$ is as follows:  
 	\begin{align}
 	\label{pool}
 	\mathbf{z}_{c}^{3\times 3}=\sum_{x=1}^{H}\sum_{y=1}^{W} \mathbf{X}^{3\times 3}_{x,y,c}*\mathbf{S}^{3\times 3}_{x,y,c} 
 	\end{align}
 	Here, $*$ denotes element multiplication. Based on the above framework, $\mathbf{z}^{5\times 5}\in \mathbb{R}^{1\times 1\times C}$ can also be obtained by input feature map $\mathbf{X}^{5\times 5}\in \mathbb{R}^{H\times W\times C}$.
 	
 	\textbf{\emph{channel attention:}} To maintain the integrity of information, we concatenate $\mathbf{z}^{3\times 3}$ and $\mathbf{z}^{5\times 5}$ as the input. Then it is transformed to a hidden representation $\mathbf{h}\in \mathbb{R}^{1\times 1\times {2*C}/{r}^{u}}$, which is always a compact feature map by setting a reduction ratio $r^{u}$ for better efficiency. This is achieved by a fully connected (FC) layer with non-linearity:  
 	\begin{align}
 	\label{hi}
 	\mathbf{h}=\tanh(BN(\mathbf{W}[\mathbf{z}^{3\times 3}\parallel \mathbf{z}^{5\times 5}])+\mathbf{b})
 	\end{align}
 	Where $BN$ is the batch normalization, $\mathbf{W}\in \mathbb{R}^{2*C \times {2*C}/{r}^{u}}$ and $\mathbf{b}\in \mathbb{R}^{{2*C}/{r}}$ denotes the weights and biases of FC layer.
 	
 	It is noteworthy that we adopt $\tanh$ rather than ReLU as our non-linearity function. For the one side, ReLU inevitably destroys feature representational power especially in low-dimensional space to a great extent, while $\tanh$ preserves information by a smoother way. For the other side, although it is widely known that $\tanh$ is more prone to cause gradient vanish as the increasing depth of CNN, this problem could not occur in our HCGNets because of the hybrid connectivity that can significantly strength the gradient back-propagation. Experimental evidence also proves that $\tanh$ is more effective than ReLU in our HCGNets. 
 	
 	Two-branch FC layers act on fusion representation $\mathbf{h}$ to generate two intermediate channel attention maps $\tilde{\mathbf{u}}^{3\times 3}\in \mathbb{R}^{1\times 1\times C}$ and $\tilde{\mathbf{u}}^{5\times 5}\in \mathbb{R}^{1\times 1\times C}$:
 	\begin{align}
 	\label{m}
 	\tilde{\mathbf{u}}^{3\times 3}=\mathbf{W}^{3 \times 3}\mathbf{h}+\mathbf{b}^{3\times 3},
 	\tilde{\mathbf{u}}^{5\times 5}=\mathbf{W}^{5 \times 5}\mathbf{h}+\mathbf{b}^{3\times 3}
 	\end{align}
 	Where $\mathbf{W}^{3 \times 3}$,$\mathbf{W}^{5 \times 5}\in  \mathbb{R}^{{2*C/r^{u}}\times {C}}$ and $\mathbf{b}^{3 \times 3}$,$\mathbf{b}^{5 \times 5}\in  \mathbb{R}^{C}$ denotes the weights and biases of two FC layers. Then a simple softmax function conducts a normalization between $\tilde{\mathbf{u}}^{3\times 3}$ and $\tilde{\mathbf{u}}^{5\times 5}$ to produce the two final channel attention maps $\mathbf{u}^{3\times 3}\in \mathbb{R}^{1\times 1\times C}$ and $\mathbf{u}^{5\times 5}\in \mathbb{R}^{1\times 1\times C}$:
 	
 	\begin{align}
 	\label{update}
 	\mathbf{u}^{3\times 3}=\frac{e^{\tilde{\mathbf{u}}^{3\times 3}}}{e^{\tilde{\mathbf{u}}^{3\times 3}}+e^{\tilde{\mathbf{u}}^{5\times 5}}},\mathbf{u}^{5\times 5}=\frac{e^{\tilde{\mathbf{u}}^{5\times 5}}}{e^{\tilde{\mathbf{u}}^{3\times 3}}+e^{\tilde{\mathbf{u}}^{5\times 5}}}
 	\end{align}
 	$\mathbf{u}^{3\times 3}$ and $\mathbf{u}^{5\times 5}$ can be regarded as the proportions of aggregating multi-scale global context features. Weighted fusion of $\mathbf{z}^{3\times 3}$ and $\mathbf{z}^{5\times 5}$ is the output of update gate:
 	\begin{align}
 	\label{fusion}
 	\mathbf{v}_{c}=\mathbf{u}_{c}^{3\times 3}\cdot \mathbf{z}_{c}^{3\times 3}+\mathbf{u}_{c}^{5\times 5}\cdot \mathbf{z}_{c}^{5\times 5}, \mathbf{u}_{c}^{3\times 3}+\mathbf{u}_{c}^{5\times 5}=1
 	\end{align}
 	Where $\mathbf{v}_{c}$ is the $c$-th channel of the output $\mathbf{v}\in \mathbb{R}^{1\times 1\times C}$.
 	
 	\subsubsection{Forget Gate.}
 	To decay the reused feature map by channel-wise weights, we locate a forget gate (see Figure \ref{Forget}) on the residual connection before information fusion. It can also be sequentially summarized for 3 stages: spatial attention, pooling and channel attention. 
 	
 	\textbf{\emph{spatial attention and pooling:}} For the given feature map $\mathbf{X}^{'}\in \mathbb{R}^{H\times W\times C}$, we perform the global attention pooling as same as update gate, thus a channel descriptor $\mathbf{z}^{f}\in \mathbb{R}^{1\times 1\times C} $ can be obtained. 
 	
 	\textbf{\emph{channel attention:}} To meet the requirement of weighted decay for each channel, the final output of each channel weight should be within $(0,1)$, thus we refer SE block, which stacks two continuous FC layers as a bottleneck and is ended by sigmoid function. Different from SE block, we insert a batch normalization layer for easing optimization and replace ReLU with $\tanh$ as our non-linearity. In short, the sequent transformations are as follows for the input $\mathbf{z}^{f}$:
 	\begin{align}
 	\label{fff}
 	\mathbf{f}=\sigma(\mathbf{W}_{2}^{f}(\tanh (BN(\mathbf{W}_{1}^{f}\mathbf{z}^{f}+\mathbf{b}^{f}_{1})))+\mathbf{b}^{f}_{2})
 	\end{align}
 	Where $\sigma$ is the sigmoid function, $\mathbf{W}_{1}^{f}\in \mathbb{R}^{C\times C/r^{f}}$, $\mathbf{b}_{1}^{f}\in \mathbb{R}^{C/r^{f}}$, $\mathbf{W}_{2}^{f}\in \mathbb{R}^{C/r^{f}\times C}$ and $\mathbf{b}_{2}^{f}\in \mathbb{R}^{C}$. $r^{f}$ is the bottleneck ratio and $\mathbf{f}\in \mathbb{R}^{1\times 1\times C}$ is the final channel attention map.
 	
 	\subsubsection{Information Fusion.} For any given feature map entering SMG module, squeeze cell firstly condenses it to a compact feature map denoted by $\mathbf{X}^{'}$. Then $\mathbf{X}^{'}$ enters multi-scale excitation cell and generate two-branch outputs $\mathbf{X}^{3\times 3}$ and $\mathbf{X}^{5\times 5}$ by 3$\times$3 and 5$\times$5 DWConvs, respectively. Since then, $\mathbf{X}^{'}$ can be regarded as the reused features, while $\mathbf{X}^{3\times 3}$ and $\mathbf{X}^{5\times 5}$ are the newly extracted features. An update gate integrates $\mathbf{X}^{3\times 3}$ and $\mathbf{X}^{5\times 5}$ to model a global context feature map $\mathbf{v}$ and we aggregate it to the decayed $\mathbf{X}^{'}$ of each spatial position by addition to build the final output $\mathbf{O}\in \mathbb{R}^{H\times W\times C}$. It can be observed that we maintain the magnitude of new features unchanged while decaying reused features, which can facilitate the effective feature exploration and retain the capability of feature re-exploitation to some extent.
 	\begin{figure}[tbp]  
 		\centering  
 		\includegraphics[width=.95\columnwidth]{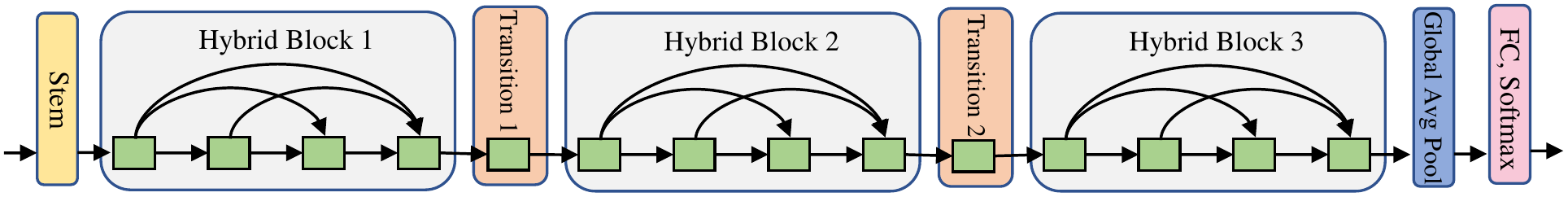}
 		\caption{A HCGNet with three hybrid blocks, where each green box denotes SMG module.}  
 		\label{macro}
 	\end{figure}
 
 	\subsection{Macro-architecture.} As shown in Figure \ref{macro}, at the beginning of HCGNet is a stem, which is a composite function to process the initial input images. Then multiple hybrid blocks are stacked with various spatial stage. Between two adjacent hybrid blocks, we adopt a transition layer to perform down-sampling and connectivity truncation. After the final hybrid block, a global average pooling attached with a softmax classifier calculates the probabilities of various categories.
 	
 	\begin{table}[tbp]
 		\centering
 		\caption{HCGNet-B network architecture for ImageNet classification. Each row describes the stage, modules information and input resolution (IR).} 
 		
 		\begin{tabular}{ccc}  
 			
 			\toprule
 			Stage&Module&IR \\
 			\midrule
 			\multirow{2}{*}{Stem}&[3$\times$3 Conv-BN-ReLU]$\times$3&224$\times$224\\
 			&3$\times$3 max pool&112$\times$112\\
 			\midrule
 			Hybrid Block&SMG$\times$3 ($k=32$)&56$\times$56\\
 			Transition&SMG$\times$1&56$\times$56\\
 			Hybrid Block&SMG$\times$6 ($k=48$)&28$\times$28\\
 			Transition&SMG$\times$1&28$\times$28\\
 			Hybrid Block&SMG$\times$12 ($k=64$)&14$\times$14\\
 			Transition&SMG$\times$1&14$\times$14\\
 			Hybrid Block&SMG$\times$8 ($k=96$)&7$\times$7\\
 			\midrule
 			\multirow{2}{*}{Classification}&global average pool&1$\times$1\\
 			&1000D FC, softmax&-\\
 			\bottomrule
 		\end{tabular}
 		\label{arch}
 	\end{table}
 	
 	Both hybrid block and transition layer adopt SMG modules but with different hyperparameter settings. We only stack one SMG module to build each transition layer and a compression factor $\theta=0.5$ is utilized to reduce the number of channels, i.e, $C=\theta \tilde{C}$. For each SMG module, we set $g=4$, $\alpha=4$ and $r^{u}=r^{f}=2$ in hybrid blocks, as well as set $g=1$, $\alpha=1.5$, $S=2$ and $r^{u}=r^{f}=4$ in transition layers. Note that we apply the standard convolutions in transition layers for best capability of feature extraction and group convolutions in hybrid blocks for better trade-off between efficiency and performance. Compared with the hybrid block, we set less multiplier $\alpha$ and larger reduction ratios $r^{u}$ and $r^{f}$ for better efficiency due to the more channels of feature maps in transition layers. 
 	
 	Specifically, we construct several networks to act on the image classification across the CIFAR and ImageNet datasets. For CIFAR, we adopt a 3$\times$3 standard convolution with stride 1 as the stem that the number of output channels is twice the growth rate of the first hybrid block. And we build three networks with various model specifications: HCGNet-(8,8,8)-($k$=12,24,36)(A1), HCGNet-(8,8,8)-($k$=24,36,64)(A2) and HCGNet-(12,12,12)-($k$=36,48,80)(A3). Formally, the first $m$-tuple indicates that there are $m$ hybrid blocks, where each figure denotes the number of SMG modules in the corresponding hybrid block. The second $m$-tuple denotes $m$ growth rates of $m$ hybrid blocks, respectively. For ImageNet, the stem consists of three contiguous 3$\times$3 Conv-BN-ReLU layers (stride 2 for the first layer) with 32, 32, 64 output channels, and attached by a 3$\times$3 max pooling with stride 2. We construct two networks: HCGNet-(3,6,12,8)-($k$=32,48,64,96)(B, as Table \ref{arch}) and HCGNet-(6,12,18,14)-($k$=48,56,72,112)(C).
 	
 	\section{Experiments}
 	\subsection{Experiments on CIFAR}
 	\begin{table}[tbp]
 		\centering 
 		\caption{Comparisons of our HCGNets against state-of-the-art  networks about test error rates (\%) across CIFAR-10 and CIFAR-100 datasets. Note that the first and second blocks contain human-designed and auto-searched architectures, respectively.}
 		\label{table1}  
 		\begin{tabular}{llccc}  
 			\toprule
 			Model&Params&FLOPs&C-10&C-100 \\  
 			\midrule
 			CondenseNet-182&4.2M&0.5G&3.76&18.47 \\
 			SparseNet-BC&16.7M&-&4.10&18.22\\
 		AOGNet&24.8M&3.7G&3.27&16.63\\
 			LogDenseNetV2 &19.0M&11.1G&3.75&18.80 \\
 			Wide ResNet-28&36.5M&5.2G&4.17&20.50\\
 			ResNeXt-29+SK &27.7M&-&3.47&17.33 \\
 			Res2NeXt-29 &36.9M&-&-&16.56 \\
 			DenseNet-BC-190 &25.6M&9.4G&3.46&17.18 \\
 			DPN-28-10&47.8M&-&3.65&20.23 \\
 			MixNet-190&48.5M&17.3G&3.13&16.96 \\
 			
 			\midrule
 			PNASNet&3.2M&-&3.41&19.53 \\
 			NASNet-A&3.3M&-&3.41&19.70\\
 			ENASNet&4.6M&-&3.54&19.43\\
 			AmoebaNet-A&4.6M&-&3.34&-\\
 			AmoebaNet-B&34.9M&-&2.98&17.66\\
 			NASNet-A &50.9M&-&-&16.03\\
 			ENASNet&52.7M&-&-&16.44\\
 			PNASNet&53.0M&-&-&16.70\\
 			\midrule
 			
 			HCGNet-A1&1.1M&0.2G&3.15&18.13\\
 			HCGNet-A2&3.1M&0.5G&2.29&16.54\\
 			HCGNet-A3&11.4M&2.0G&\textbf{2.14}&\textbf{15.96} \\
 			\bottomrule
 		\end{tabular}
 		
 	\end{table}
 	
 	\subsubsection{Dataset and training details.} Both CIFAR-10 and CIFAR-100 datasets comprise 50k training images and 10k test images corresponding to 10 and 100 classes, respectively. We apply a standard data augmentation following \citet{huang2017densely}. We employ a stochastic gradient descent (SGD) optimizer with momentum 0.9 and batch size 128. Training is regularized by weight decay $1\times 10^{-4}$ and mixup with $\alpha=1$ \cite{zhang2017mixup}. For HCGNet-A1, we train it for 1270 epochs by SGDR \cite{loshchilov2016sgdr} learning rate curve with initial learning rate 0.1, $T_{0}=10$, $T_{mul}=2$. For HCGNet-A2 and A3, we train them for 1260 epochs including two continuous 630 epochs, each of them is a SGDR learning rate curve with initial learning rate 0.1, $T_{0}=10$, $T_{mul}=2$.

 \subsubsection{Comparisons with Human-designed Networks.} Quantitatively in Table \ref{table1}, DenseNet-190 has 31 modules in each dense block, while HCGNet-A2 only has 8 modules in each hybrid block thus reduces 93\% redundancy but with substantial accuracy gains. Moreover, HCGNet-A2 significantly outperforms other sparsification variants, such as LogDenseNet, SparseNet and CondenseNet, which indicate that our optimization of DenseNet is more effective than sparsification method. HCGNet-A2 using $16\times $ fewer parameters surpasses MixNet-190, which represents the most general form of ResNet and DenseNet. It also uses $8\times $ fewer parameters but obtains better results than concurrent AOGNet, which is the state-of-the-art human network by hierarchical and compositional feature aggregation. Consequently, our nested aggregation is the best method among other combinations and variants of ResNet and DenseNet. 

\begin{table}[tbp]
	\centering
	\caption{Comparisons of our HCGNets against SOTA networks about Top-1 and Top-5 error rates (\%) on ImageNet.}
	\label{imagenet}  
	\begin{tabular}{lcccc}  
		\toprule
		Model&Params&FLOPs&T-1&T-5 \\  
		\midrule
		MixNet-105&11.2M&5.0G&23.3&6.7\\
		MixNet-121&21.9M&8.3G&21.9&5.9\\
		MixNet-141&41.1M&13.1G&20.4&5.3\\	 
		\midrule
		DPN-68 &12.8M&2.5G&23.6&6.9 \\
		DPN-92 &38.0M&6.5G&20.7&5.4 \\
		DPN-98 &61.6M&11.7G&20.2&5.2 \\
		\midrule
		DenseNet-169 &14.2M&3.5G&23.8&6.9 \\
		DenseNet-201 &20.0M&4.4G&22.6&6.3 \\
		DenseNet-264 &33.4M&6.0G&22.2&6.1 \\
		\midrule
		SparseNet-201 &14.9M&9.2G&22.7&- \\
		\midrule
		ResNet-50&25.6M&3.9G&24.6&7.5 \\
		ResNet-50+SE&28.1M&3.9G&23.1&6.7 \\
		ResNet-50+CBAM&28.1M&3.9G&22.7&6.3 \\
		ResNet-101&44.6M&7.6G&23.4&6.9 \\
		ResNet-101+SE&49.3M&7.6G&22.4&6.2 \\
		ResNet-101+CBAM&49.3M&7.6G&21.5&5.7 \\
		\midrule
		ResNeXt-50&25.0M&3.8G&22.9&6.5 \\
		ResNeXt-50+SE&27.6M&3.8G&21.9&6.0 \\
		ResNeXt-101&44.2M&7.5G&21.5&5.8 \\
		ResNeXt-101+SE&49.0M&7.5G&21.2&5.7 \\
		ResNeXt-101+SK&48.9M&8.5G&20.2&- \\
		\midrule
		WideResNet-18&45.6M&6.7G&25.6&8.2 \\
		WideResNet-18+SE&46.0M&6.7G&24.9&7.7 \\
		\midrule
		AOGNet-12M&11.9M&2.4G&22.3&6.1 \\
		AOGNet-40M&40.3M&8.9G&19.8&4.9 \\
		\midrule
		HCGNet-B&12.9M&2.0G&21.5&5.8 \\
		HCGNet-C&42.2M&7.1G&\textbf{19.5}&\textbf{4.8} \\
		\bottomrule
	\end{tabular}
\end{table}

 \begin{table*}[t]
	\centering 
	\caption{Comparisons of HCGNet-B against other backbones on the Mask-RCNN system \cite{he2017mask}. } 
	\label{gty}
	\begin{tabular}{lcccccccc}  
		\toprule
		Backbone &Params&FLOPs&AP$^{bb}_{50:95}$&AP$^{bb}_{50}$&AP$^{bb}_{75}$&AP$^{m}_{50:95}$&AP$^{m}_{50}$&AP$^{m}_{75}$ \\
	\midrule
		ResNet-50-FPN&44.2M&275.6G&37.3&59.9&40.2&34.2&55.9&36.2 \\ 
		AOGNet-12M-FPN&31.2M&-&38.0&59.8&\textbf{41.3}&34.6&56.6&36.4 \\ 
	\midrule
		HCGNet-B-FPN&32.1M&230.4G&\textbf{38.3}&\textbf{60.6}&\textbf{41.3}&\textbf{35.2}&\textbf{57.5}&\textbf{37.1} \\
	\bottomrule
	\end{tabular}
\end{table*}
 \subsubsection{Comparisons with auto-searched Networks.} Notably, Our HCGNets are also more efficient than auto-searched networks. Compared with other networks with small setting, HCGNet-A2 achieves around 1\% and 3\% reductions on CIFAR-10 and CIFAR-100 error rates, respectively. Moreover, it is noteworthy that HCGNet-A1 can also obtain superior performance with unprecedent efficiency. For large setting, HCGNet-A3 achieves the best results with least complexity. Somewhat surprisingly, HCGNet-A3 can outperform the most competitive NASNet-A with only 22\% parameters.

 	\subsection{Experiments on ImageNet 2012}
 		
 	\begin{figure}[tbp]
 		\centering  
 		\includegraphics[width=.95\columnwidth]{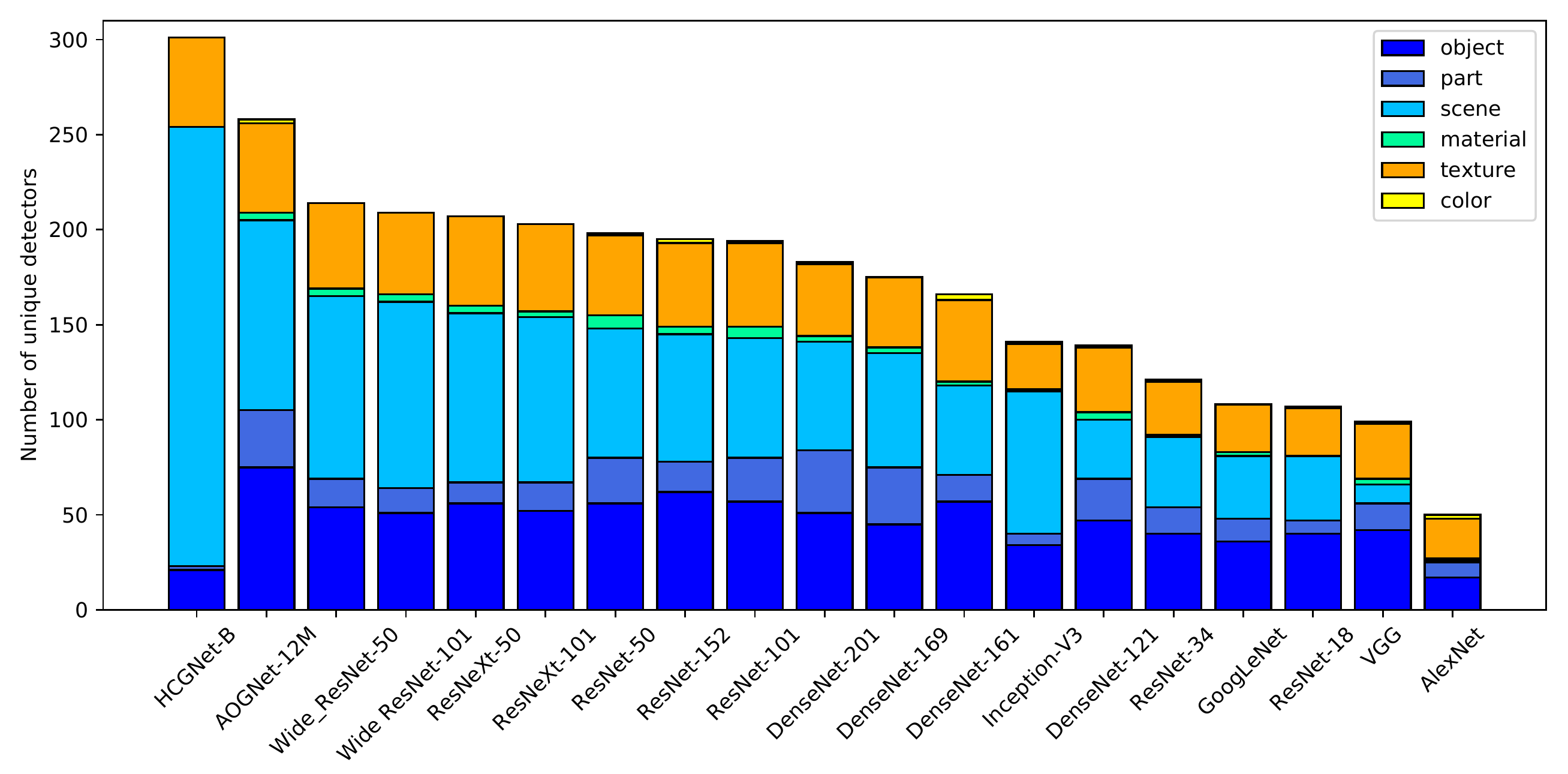}
 		\caption{Comparisons of interpretability by network dissection \cite{bau2017network} among popular models based on ImageNet pretrained models.}  
 		\label{interpretability}
 	\end{figure}
 	\subsubsection{Dataset and Training Details.} ImageNet 2012 dataset comprises 1.2 million training images and 50k validation images corresponding to 1000 classes. We employ the data augmentation following \citet{huang2017densely}. Final error rates are reported by single-crop with size $224\times 224$ at test time on the validation set. We employ synchronous SGD with momentum 0.9 and batch size 256. Training is regularized by weight decay $4\times 10^{-5}$, label smoothing with $\epsilon=0.1$ \cite{szegedy2016rethinking}, mixup with $\alpha=0.4$ and dropout \cite{srivastava2014dropout} with rate 0.1 before the final FC layer. All networks are trained for 630 epochs by SGDR learning rate curve with initial learning rate 0.1, $T_{0}=10$, $T_{mul}=2$.
 	\subsubsection{Comparisons with popular networks.} As shown in Table \ref{imagenet}, our HCGNets perform the best among all other models with less or comparable complexity in terms of top-1 and top-5 error rates. It is noteworthy that DenseNet-169 stacks 4 dense blocks with 6,12,32,32 modules, while HCGNet-B utilizes shallower design with 3,6,12,8 modules for 4 hybrid blocks, thus reducing 88\% redundancy but obtaining above absolute 2.3\% gain of performance. Furthermore, HCGNets yield significantly better results than the families of DenseNet, MixNet and DPN under comparable complexity. Remarkably, using considerable $4.6\times$ fewer FLOPs, HCGNet-B can also surpass SparseNet-201, which is the state-of-the-art variant of DenseNet. The family of HCGNet can consistently obtain better performance than the families of ResNet, ResNeXt, WideResNet and their attention-based variants, which represent the widely applied
 	models in practice. Predominately, HCGNets outperform previous
 	SOTA AOGNets across various model specifications, which show
 	the superiority of our design.

 	\subsubsection{Model Interpretability} We quantify the interpretability by network dissection, which compares the number of unique detectors in the final convolutional layer. Figure \ref{interpretability} shows that HCGNet-B obtains the overall highest score with least complexity, which shows that the designs of hybrid connectivity and SMG module can not only achieve the best accuracy, but also generate the best latent representations.
 	\begin{figure}[tbp]
 		\centering  
 		\includegraphics[width=.95\columnwidth]{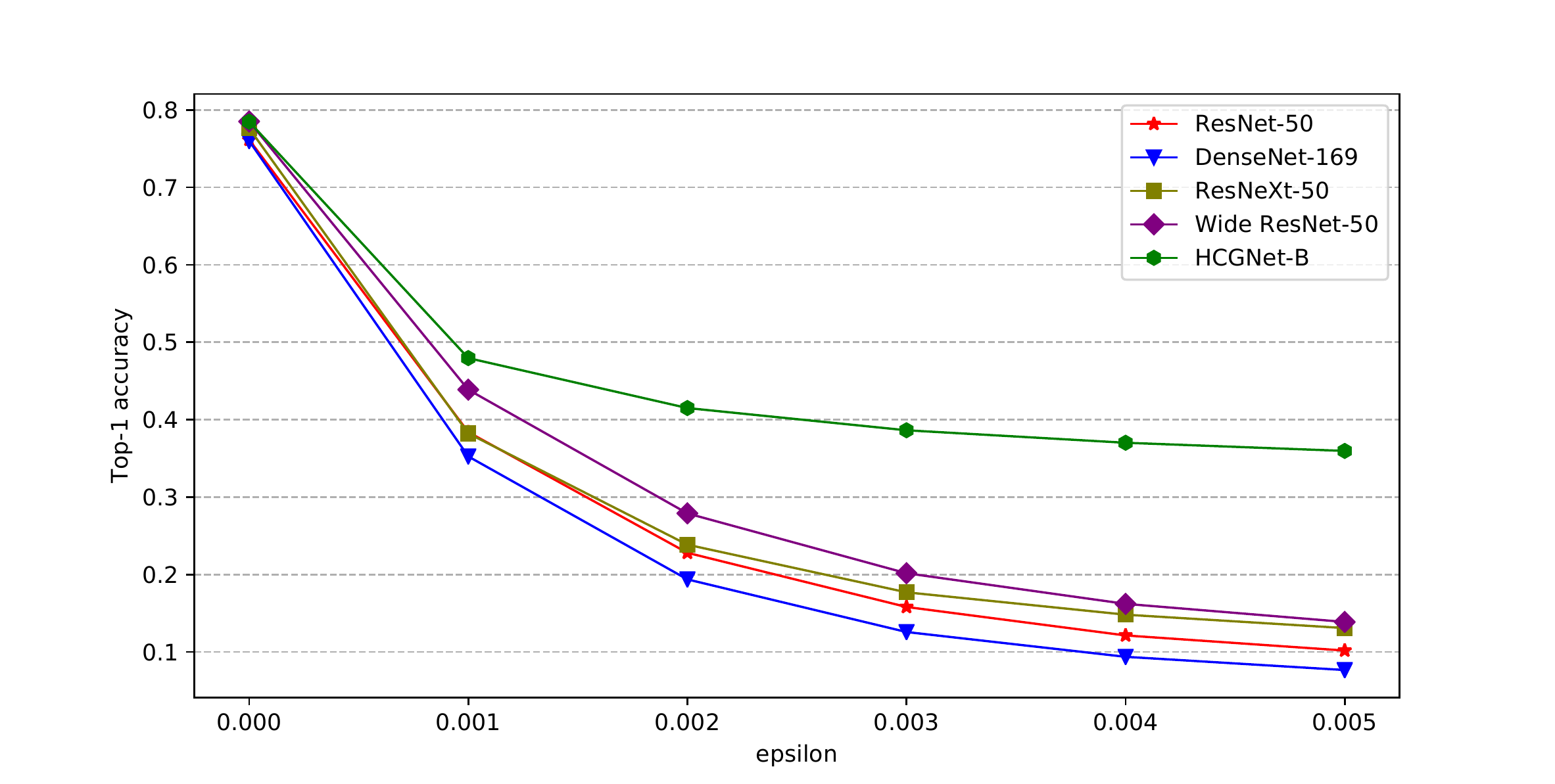}
 		\caption{Comparisons of adversarial robustness by FGSM \cite{attack} attack.}  
 		\label{attack}
 	\end{figure}
 	\subsubsection{Adversarial Robustness}
 	We attack HCGNet-B by popular FGSM across various perturbation energies $\epsilon$ to test the adversarial robustness against widely applied models, results of which are shown in Figure \ref{attack}. HCGNet-B has a more remarkable robustness than other models in adversarial defense, especially the perturbation is relatively high.
 	
 	\subsection{Object Detection and Instance Segmentation}
 	To show the transferability, we experiment HCGNet-B pretrained on ImageNet as a backbone on the Mask-RCNN system to implement object detection and instance segmentation tasks. We use COCO train2017 set to finetune the HCGNet-B by the 1x training schedule, and evaluate the performance on COCO val2017 set. We report the results by standard COCO metrics of Average Precision (AP), i.e, AP$_{50:95}$, AP$_{50}$, and AP$_{75}$ for bounding box detection (AP$^{bb}$) and instance segmentation (AP$^{m}$)  in Table \ref{gty}. The results show that HCGNet-B can learn better features than SOTA ResNet and AOGNet backbones.
 	
 	\section{Conclusion.}
 	This paper develops an efficient architecture with the innovative designs of hybrid connectivity, micro-module and attention-based forget and update gates. On CIFAR and ImageNet datasets, HCGNets outperform state-of-the-art networks with less or comparable complexity. Extensive experiments based on the ImageNet pretrained model further show the remarkable interpretability, robustness for recognition and transferability for detection. We hope our HCGNets may inspire the future study of architectural design and search.
 	
 	\bibliographystyle{aaai}
 	\bibliography{5011-aaai.bib}

\begin{thebibliography}{}

\bibitem[\protect\citeauthoryear{Bau \bgroup et al\mbox.\egroup
  }{2017}]{bau2017network}
Bau, D.; Zhou, B.; Khosla, A.; Oliva, A.; and Torralba, A.
\newblock 2017.
\newblock Network dissection: Quantifying interpretability of deep visual
  representations.
\newblock In {\em Proceedings of the IEEE Conference on Computer Vision and
  Pattern Recognition},  6541--6549.

\bibitem[\protect\citeauthoryear{Cao \bgroup et al\mbox.\egroup
  }{2019}]{cao2019gcnet}
Cao, Y.; Xu, J.; Lin, S.; Wei, F.; and Hu, H.
\newblock 2019.
\newblock Gcnet: Non-local networks meet squeeze-excitation networks and
  beyond.
\newblock {\em arXiv preprint arXiv:1904.11492}.

\bibitem[\protect\citeauthoryear{Chen \bgroup et al\mbox.\egroup
  }{2017}]{chen2017dual}
Chen, Y.; Li, J.; Xiao, H.; Jin, X.; Yan, S.; and Feng, J.
\newblock 2017.
\newblock Dual path networks.
\newblock In {\em Advances in Neural Information Processing Systems},
  4467--4475.

\bibitem[\protect\citeauthoryear{Deng \bgroup et al\mbox.\egroup
  }{2009}]{deng2009imagenet}
Deng, J.; Dong, W.; Socher, R.; Li, L.-J.; Li, K.; and Fei-Fei, L.
\newblock 2009.
\newblock Imagenet: A large-scale hierarchical image database.
\newblock In {\em IEEE conference on computer vision and pattern recognition},
  248--255.

\bibitem[\protect\citeauthoryear{Gao \bgroup et al\mbox.\egroup
  }{2019}]{gao2019res2net}
Gao, S.-H.; Cheng, M.-M.; Zhao, K.; Zhang, X.-Y.; Yang, M.-H.; and Torr, P.
\newblock 2019.
\newblock Res2net: A new multi-scale backbone architecture.
\newblock {\em arXiv preprint arXiv:1904.01169}.

\bibitem[\protect\citeauthoryear{Goodfellow, Shlens, and
  Szegedy}{2015}]{attack}
Goodfellow, I.~J.; Shlens, J.; and Szegedy, C.
\newblock 2015.
\newblock Explaining and harnessing adversarial examples.
\newblock In {\em 3rd International Conference on Learning Representations}.

\bibitem[\protect\citeauthoryear{He \bgroup et al\mbox.\egroup
  }{2016}]{he2016deep}
He, K.; Zhang, X.; Ren, S.; and Sun, J.
\newblock 2016.
\newblock Deep residual learning for image recognition.
\newblock In {\em Proceedings of the IEEE conference on computer vision and
  pattern recognition},  770--778.

\bibitem[\protect\citeauthoryear{He \bgroup et al\mbox.\egroup
  }{2017}]{he2017mask}
He, K.; Gkioxari, G.; Doll{\'a}r, P.; and Girshick, R.
\newblock 2017.
\newblock Mask r-cnn.
\newblock In {\em Proceedings of the IEEE international conference on computer
  vision},  2961--2969.

\bibitem[\protect\citeauthoryear{Hu \bgroup et al\mbox.\egroup
  }{2017}]{hu2017log}
Hu, H.; Dey, D.; Del~Giorno, A.; Hebert, M.; and Bagnell, J.~A.
\newblock 2017.
\newblock Log-densenet: How to sparsify a densenet.
\newblock {\em arXiv preprint arXiv:1711.00002}.

\bibitem[\protect\citeauthoryear{Hu, Shen, and Sun}{2018}]{hu2018squeeze}
Hu, J.; Shen, L.; and Sun, G.
\newblock 2018.
\newblock Squeeze-and-excitation networks.
\newblock In {\em Proceedings of the IEEE conference on computer vision and
  pattern recognition},  7132--7141.

\bibitem[\protect\citeauthoryear{Huang \bgroup et al\mbox.\egroup
  }{2017}]{huang2017densely}
Huang, G.; Liu, Z.; Van Der~Maaten, L.; and Weinberger, K.~Q.
\newblock 2017.
\newblock Densely connected convolutional networks.
\newblock In {\em Proceedings of the IEEE conference on computer vision and
  pattern recognition},  4700--4708.

\bibitem[\protect\citeauthoryear{Huang \bgroup et al\mbox.\egroup
  }{2018}]{huang2018condensenet}
Huang, G.; Liu, S.; Van~der Maaten, L.; and Weinberger, K.~Q.
\newblock 2018.
\newblock Condensenet: An efficient densenet using learned group convolutions.
\newblock In {\em Proceedings of the IEEE Conference on Computer Vision and
  Pattern Recognition},  2752--2761.

\bibitem[\protect\citeauthoryear{Krizhevsky and
  Hinton}{2009}]{krizhevsky2009learning}
Krizhevsky, A., and Hinton, G.
\newblock 2009.
\newblock Learning multiple layers of features from tiny images.
\newblock Technical report, Citeseer.

\bibitem[\protect\citeauthoryear{Li \bgroup et al\mbox.\egroup
  }{2019}]{li2019selective}
Li, X.; Wang, W.; Hu, X.; and Yang, J.
\newblock 2019.
\newblock Selective kernel networks.
\newblock In {\em Proceedings of the IEEE conference on computer vision and
  pattern recognition}.

\bibitem[\protect\citeauthoryear{Li, Song, and Wu}{2019}]{AOGNets}
Li, X.; Song, X.; and Wu, T.
\newblock 2019.
\newblock Aognets: Compositional grammatical architectures for deep learning.
\newblock In {\em Proceedings of the IEEE conference on computer vision and
  pattern recognition}.

\bibitem[\protect\citeauthoryear{Liu \bgroup et al\mbox.\egroup
  }{2018}]{liu2018progressive}
Liu, C.; Zoph, B.; Neumann, M.; Shlens, J.; Hua, W.; Li, L.-J.; Fei-Fei, L.;
  Yuille, A.; Huang, J.; and Murphy, K.
\newblock 2018.
\newblock Progressive neural architecture search.
\newblock In {\em Proceedings of the European Conference on Computer Vision},
  19--34.

\bibitem[\protect\citeauthoryear{Loshchilov and
  Hutter}{2016}]{loshchilov2016sgdr}
Loshchilov, I., and Hutter, F.
\newblock 2016.
\newblock Sgdr: Stochastic gradient descent with warm restarts.
\newblock {\em arXiv preprint arXiv:1608.03983}.

\bibitem[\protect\citeauthoryear{Real \bgroup et al\mbox.\egroup
  }{2019}]{real2019regularized}
Real, E.; Aggarwal, A.; Huang, Y.; and Le, Q.~V.
\newblock 2019.
\newblock Regularized evolution for image classifier architecture search.
\newblock In {\em Proceedings of the AAAI Conference on Artificial
  Intelligence}, volume~33,  4780--4789.

\bibitem[\protect\citeauthoryear{Srivastava \bgroup et al\mbox.\egroup
  }{2014}]{srivastava2014dropout}
Srivastava, N.; Hinton, G.; Krizhevsky, A.; Sutskever, I.; and Salakhutdinov,
  R.
\newblock 2014.
\newblock Dropout: a simple way to prevent neural networks from overfitting.
\newblock {\em The journal of machine learning research} 15(1):1929--1958.

\bibitem[\protect\citeauthoryear{Szegedy \bgroup et al\mbox.\egroup
  }{2016}]{szegedy2016rethinking}
Szegedy, C.; Vanhoucke, V.; Ioffe, S.; Shlens, J.; and Wojna, Z.
\newblock 2016.
\newblock Rethinking the inception architecture for computer vision.
\newblock In {\em Proceedings of the IEEE conference on computer vision and
  pattern recognition},  2818--2826.

\bibitem[\protect\citeauthoryear{Szegedy \bgroup et al\mbox.\egroup
  }{2017}]{szegedy2017inception}
Szegedy, C.; Ioffe, S.; Vanhoucke, V.; and Alemi, A.~A.
\newblock 2017.
\newblock Inception-v4, inception-resnet and the impact of residual connections
  on learning.
\newblock In {\em Thirty-First AAAI Conference on Artificial Intelligence},
  4278--4284.

\bibitem[\protect\citeauthoryear{Wang \bgroup et al\mbox.\egroup
  }{2017}]{wang2017residual}
Wang, F.; Jiang, M.; Qian, C.; Yang, S.; Li, C.; Zhang, H.; Wang, X.; and Tang,
  X.
\newblock 2017.
\newblock Residual attention network for image classification.
\newblock In {\em Proceedings of the IEEE Conference on Computer Vision and
  Pattern Recognition},  3156--3164.

\bibitem[\protect\citeauthoryear{Wang \bgroup et al\mbox.\egroup
  }{2018a}]{wang2018mixed}
Wang, W.; Li, X.; Yang, J.; and Lu, T.
\newblock 2018a.
\newblock Mixed link networks.
\newblock {\em arXiv preprint arXiv:1802.01808}.

\bibitem[\protect\citeauthoryear{Wang \bgroup et al\mbox.\egroup
  }{2018b}]{wang2018non}
Wang, X.; Girshick, R.; Gupta, A.; and He, K.
\newblock 2018b.
\newblock Non-local neural networks.
\newblock In {\em Proceedings of the IEEE Conference on Computer Vision and
  Pattern Recognition},  7794--7803.

\bibitem[\protect\citeauthoryear{Woo \bgroup et al\mbox.\egroup
  }{2018}]{woo2018cbam}
Woo, S.; Park, J.; Lee, J.-Y.; and So~Kweon, I.
\newblock 2018.
\newblock Cbam: Convolutional block attention module.
\newblock In {\em Proceedings of the European Conference on Computer Vision},
  3--19.

\bibitem[\protect\citeauthoryear{Xie \bgroup et al\mbox.\egroup
  }{2017}]{xie2017aggregated}
Xie, S.; Girshick, R.; Doll{\'a}r, P.; Tu, Z.; and He, K.
\newblock 2017.
\newblock Aggregated residual transformations for deep neural networks.
\newblock In {\em Proceedings of the IEEE conference on computer vision and
  pattern recognition},  1492--1500.

\bibitem[\protect\citeauthoryear{Yu and Koltun}{2016}]{DBLP}
Yu, F., and Koltun, V.
\newblock 2016.
\newblock Multi-scale context aggregation by dilated convolutions.
\newblock In {\em International Conference on Learning Representations}.

\bibitem[\protect\citeauthoryear{Zagoruyko and Komodakis}{2016}]{Wide}
Zagoruyko, S., and Komodakis, N.
\newblock 2016.
\newblock Wide residual networks.
\newblock In {\em Proceedings of the British Machine Vision Conference}.

\bibitem[\protect\citeauthoryear{Zhang \bgroup et al\mbox.\egroup
  }{2017}]{zhang2017mixup}
Zhang, H.; Cisse, M.; Dauphin, Y.~N.; and Lopez-Paz, D.
\newblock 2017.
\newblock mixup: Beyond empirical risk minimization.
\newblock {\em arXiv preprint arXiv:1710.09412}.

\bibitem[\protect\citeauthoryear{Zhu \bgroup et al\mbox.\egroup
  }{2018}]{zhu2018sparsely}
Zhu, L.; Deng, R.; Maire, M.; Deng, Z.; Mori, G.; and Tan, P.
\newblock 2018.
\newblock Sparsely aggregated convolutional networks.
\newblock In {\em Proceedings of the European Conference on Computer Vision},
  186--201.

\bibitem[\protect\citeauthoryear{Zoph \bgroup et al\mbox.\egroup
  }{2018}]{zoph2018learning}
Zoph, B.; Vasudevan, V.; Shlens, J.; and Le, Q.~V.
\newblock 2018.
\newblock Learning transferable architectures for scalable image recognition.
\newblock In {\em Proceedings of the IEEE conference on computer vision and
  pattern recognition},  8697--8710.

\end{thebibliography}
 	
 \end{document}